\documentclass[journal]{IEEEtran}

\usepackage{times}
\usepackage{epsfig}
\usepackage{amssymb}
\usepackage{subfigure}
\usepackage{bbding}
\usepackage{xcolor}
\usepackage{threeparttable}
\usepackage{booktabs}
\usepackage{algorithm}
\usepackage{algorithmic}
\usepackage{etoolbox}
\makeatletter
\patchcmd{\@makecaption}
  {\scshape}
  {}
  {}
  {}
\makeatother

\newcommand{\eg}{\textit{e}.\textit{g}.}
\newcommand{\ie}{\textit{i}.\textit{e}.}
\newcommand{\et}{\textit{e}\textit{t} \textit{a}\textit{l}.}

\usepackage{graphicx, color, multirow, amsmath}

\RequirePackage{latexsym,amsmath,amssymb}

\ifCLASSINFOpdf
\else
  have to specify these with
\fi

\begin{document}

\title{Video Quality Assessment for Online Processing: From Spatial to Temporal Sampling}
\author{
Jiebin Yan,
Lei Wu,
Yuming Fang,~\IEEEmembership{Senior~Member,~IEEE},
Xuelin Liu,
Xue Xia,
Weide Liu

\thanks{This work was supported in part by the National Key Research and Development Program of China under Grant 2023YFE0210700, in part by the National Natural Science Foundation of China under Grants 62132006, 62461028, 62441203, 62162029, and 62311530101, in part by the Natural Science Foundation of Jiangxi Province of China under Grants 20223AEI91002 and 20232BAB202001, and in part by the project funded by China Postdoctoral Science Foundation under Grant 2024T170364. (Corresponding author: Yuming Fang).}

\thanks{Jiebin Yan, Lei Wu, Yuming Fang, Xuelin Liu, and Xue Xia are with the School of Information Technology, Jiangxi University of Finance and Economics, Nanchang 330032, Jiangxi, China. (e-mail: yanjiebin@jxufe.edu.cn;fa0001ng@e.ntu.edu.sg; yeziandkuma@qq.com).}

\thanks{W. Liu is with Harvard Medical School, Harvard University, USA (e-mail: weide001@e.ntu.edu.sg).}



}


\maketitle

\begin{abstract}
With the rapid development of multimedia processing and deep learning technologies, especially in the field of video understanding, video quality assessment (VQA) has achieved significant progress. Although researchers have moved from designing efficient video quality mapping models to various research directions, in-depth exploration of the \textit{effectiveness-efficiency trade-offs} of spatio-temporal modeling in VQA models is still less sufficient. Considering the fact that videos have highly redundant information, this paper investigates this problem from the perspective of joint spatial and temporal sampling, aiming to seek the answer to how little information we should keep at least when feeding videos into the VQA models while with acceptable performance sacrifice. To this end, we drastically sample the video's information from both spatial and temporal dimensions, and the \textit{heavily squeezed} video is then fed into a stable VQA model. Comprehensive experiments regarding joint spatial and temporal sampling are conducted on six public video quality databases, and the results demonstrate the acceptable performance of the VQA model when throwing away most of the video information. Furthermore, with the proposed joint spatial and temporal sampling strategy, we make an initial attempt to design an online VQA model, which is instantiated by as simple as possible a spatial feature extractor, a temporal feature fusion module, and a global quality regression module. Through quantitative and qualitative experiments, we verify the feasibility of online VQA model by simplifying itself and reducing input. 
\end{abstract}


\begin{IEEEkeywords}
Video quality assessment, spatial and temporal sampling, video understanding.
\end{IEEEkeywords}

\IEEEpeerreviewmaketitle

\section{Introduction}
\label{sec:intr}

\IEEEPARstart{I}{n} the past few years, the number of user-generated videos has grown exponentially, and video has become one of the important elements in our daily life. However, in the whole chain from video generation, transmission, storage, processing to display, distortion would be inevitably introduced, resulting in quality degradation~\cite{yan23a,yan22the}, and therefore accurate evaluation of video quality becomes a core problem in real applications~\cite{fang2021superpixel,le2021perceptually,wang2011applications}. Video quality assessment (VQA) has proverbially been a classic and important research topic in video understanding, and it can be achieved through both subjective and objective quality assessment. Instead of quality evaluation by human beings (\ie, subjective quality assessment, which is considered the most reliable but labor-intensive and expensive method), objective VQA aims to design computational models that can automatically and accurately predict the human perceptual quality of videos. According to the availability of reference information, VQA models could be categorized into full-reference (FR), reduced-reference (RR), and no-reference (NR) or blind (B) methods. The first two types of VQA models require full or partial reference information when being used to evaluate video quality, while the last type of VQA models do not need any reference information. In terms of practicability, BVQA methods can predict the quality of videos without access to reference information and thus are more applicable~\cite{min2024perceptual}.

\begin{figure}[]
\centering
\includegraphics[scale=0.46]{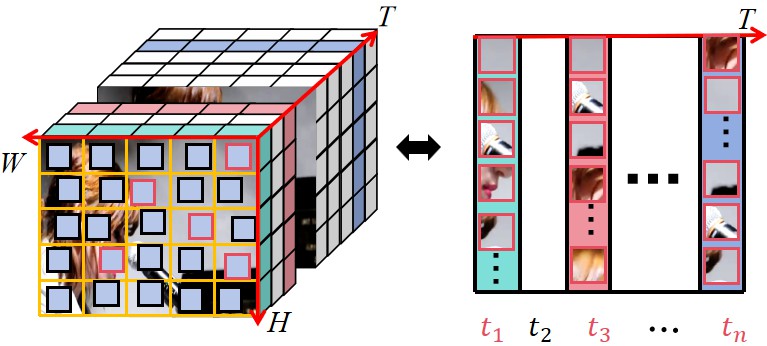}
\caption{An illustration of spatio-temporal sampling paradigm, which extracts representatives for quality prediction from stacked spatio-temporal blocks. Note that a video can be broken down into many spatio-temporal blocks by spatio-temporal grid division.}
\label{fig:video_com}
\end{figure}

Generally, in the handcrafted features dominated era, the VQA task can be decomposed in-frame (or short-duration video segment) quality measurement and frame-by-frame (or video segment) error pooling~\cite{wang2001human}, where the former part always refers to an error-sensitivity metrics such as Peak Signal to Noise Ratio (PSNR) and structural similarity (SSIM)~\cite{wang2004image}, and the latter part relies on the characteristics of the human visual system (HVS) such as motion perceptual uncertainty~\cite{wang2007video} and the recency effect~\cite{pearson1998viewer}. Currently, deep learning technology makes these two parts being formulated in a unified framework, \ie, adaptively learning spatio-temporal quality degradation patterns in an end-to-end manner. Despite the significant success, computational complexity has always been a substantial issue in designing deep learning based VQA models, which is similar and contemporaneous to video understanding~\cite{kantorov2014efficient}. A straightforward and high-gain way to reduce computational cost is to discard part of video information by resizing, cropping, sampling, and other operations, since it is common sense that videos have large redundant information in spatio-temporal dimensions. However, the widely used downsampling operations, such as resizing and random cropping, would lead to the artificial destruction of videos. For example, the resizing operation may introduce new distortion and change the original video quality, and the cropping operation can cause a mismatch between global and local quality. Moreover, it is more challenging for current VQA models to deal with long videos, which contain visual distortions possibly localized in space and time.

As revealed in our previous studies~\cite{yan2022subjective,FangTIP2023}, a limited number of sparsely sampled video frames is capable of being used to predict video quality with acceptable performance. To be more specific, Yan~\et~\cite{yan2022subjective} empirically found that modifying the original input setting of VSFA~\cite{li2019quality} to few frames (temporal coherence is lost to some extent) of free viewpoint videos can also obtain competing performance. Fang~\et~\cite{FangTIP2023} further extended this work~\cite{yan2022subjective} from one more perspective, \ie, considering the order of input frames in both training and testing phases, and found that the order of video frames has less influence on model performance~\cite{li2019quality}. Moreover, Wu~\et~\cite{wu2022fast} proposed a new concept namely fragments to represent the test video, which naturally conforms with the transform framework. Sun~\et~\cite{sun2024analysis} explored the VQA problem via designing minimalistic video quality models. However, the \textit{effectiveness-efficiency trade-offs} of spatio-temporal modeling in VQA is still not well explored, \ie, how less video information we should keep at least when feeding videos into the VQA models with acceptable performance sacrifice. To probe the answer to the above question, in this paper, we conduct comprehensive experiments on spatio-temporal modeling in VQA with a specific focus on simplicity and lightweight. Accordingly, two key elements are mainly considered, including minimal input (as shown in Figure~\ref{fig:video_com}) and a simple spatial and temporal feature extractor. Note that there is much difference between our study and these studies~\cite{yan2022subjective,FangTIP2023,wu2022fast,sun2024analysis}. First, compared to these studies~\cite{yan2022subjective,FangTIP2023}, we investigate the influence of information loss on video quality prediction from both spatial and temporal dimensions, rather than solely from temporal dimension. Second, the study~\cite{wu2022fast} pursued the efficiency by spatio-temporal grid mini-cube sampling, while its exploration on temporal sampling is still insufficient, and its computational cost is still large. Third, Sun~\et~\cite{sun2024analysis} delved into the easy database problem in VQA, however, the experimented models (\ie, various variants of the baseline) rely on computation-extensive spatial extractor and motion extractor. Apart from a detailed exploration of joint spatial and temporal sampling, we make a further attempt to design an online VQA model.

\begin{figure*}[t]
\centering
\includegraphics[scale=0.44]{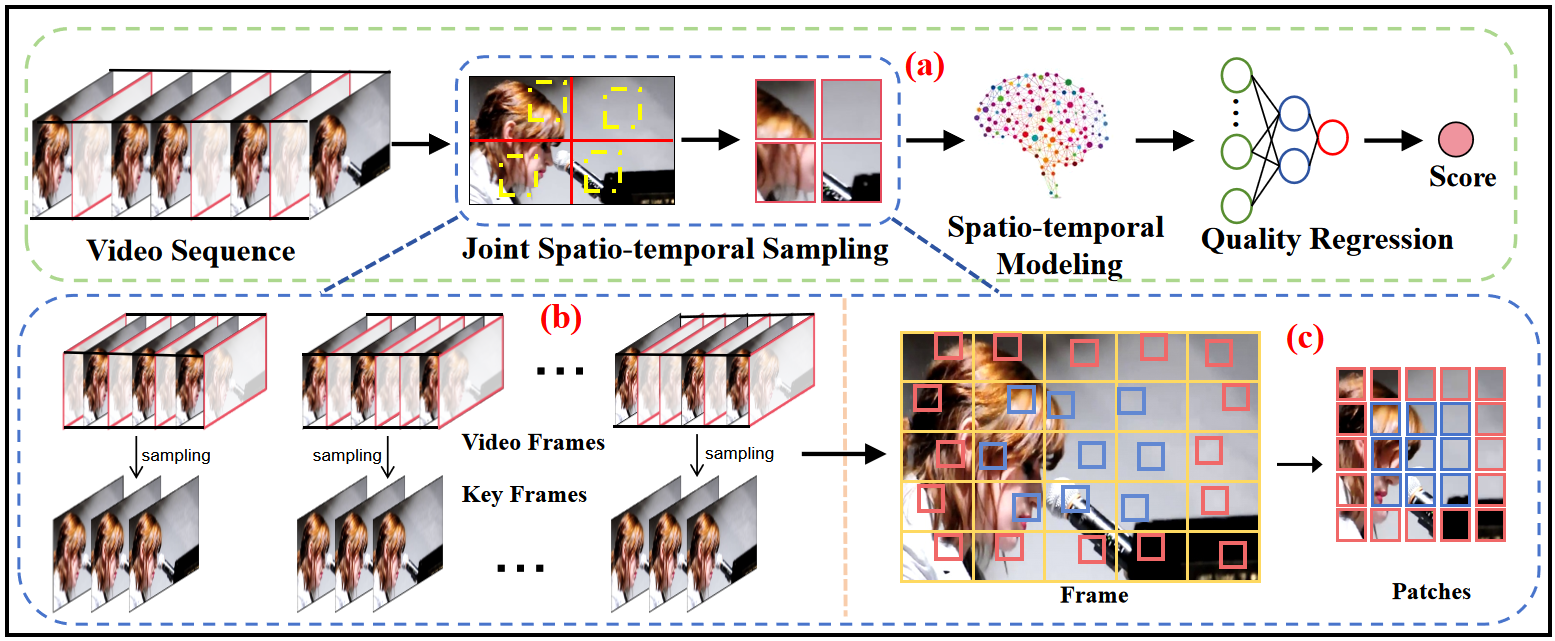}
\caption{The whole framework of this study (a). Given a video sequence, we first squeeze the input video by joint spatial and temporal sampling (b and c), and can obtain stacked spatio-temporal blocks. The squeezed video is then fed into spatio-temporal modeling module, which is instantiated by a spatial feature extractor and a temporal feature fusion module under the philosophy of minimalism. Finally, the extracted features are then used to predict video quality.}
\label{fig:motive}
\end{figure*}

In summary, our contributions are three folds:
\begin{itemize}
\item We empirically present a comprehensive study on online VQA from spatial to temporal sampling, investigating the possibility of pursuing the effectiveness-efficiency trade-offs of spatio-temporal modeling in VQA. 
\item We elaborately conduct a series of quantitative experiments via jointly integrating spatial sampling schemes and temporal sampling strategies, where the spatio-temporal modeling module is instantiated by a vanilla gated recurrent units (GRU) or transformer under the philosophy of minimalism. 
\item We take a further step to design an online VQA model named MGQA, which includes a spatial feature extractor, a temporal feature fusion module, and a global quality regression module. The average computational cost across six test video quality databases is reduced by 99.83\% compared to the stable VQA model, \ie, VSFA~\cite{li2019quality}, in terms of processed data. 

\end{itemize}

\section{Related Work}
\label{sec:rw}

In this section, we first introduce VQA models. Then, we describe studies about online video processing. 
\subsection{VQA Models}
\label{subsec:vd}

In general, BVQA methods can be classified into two categories according to the design philosophy, including knowledge-driven and data-driven VQA models. The knowledge-driven BVQA models mainly rely on handcrafted features, \eg, natural scene statistics (NSS), and use a shallow machine learning algorithm to map the features to the visual quality of videos. NSS refers to the statistical regularities of visual scenes, where the statistical discrepancy between high-quality video and test video represents the visual quality of the test video. In these types of VQA models, spatial and temporal information are crucial. For example, Saad~\et~\cite{Vbliinds} predicted the visual quality of videos by estimating spatial naturalness, spatio-temporal statistical regularity naturalness, and motion coherency. Zhu~\et~\cite{zhu2022learning} proposed a spatio-temporal interaction model by integrating spatial features and temporal motion features for evaluating video perceptual quality. Mittal~\et~\cite{VIIDEO} proposed a BVQA model named VIIDEO that exploits inherent statistical regularities of natural videos. Korhonen~\et~\cite{TLVQM} extracted a set of handcrafted spatio-temporal features to predict the quality of videos. In ChipQA~\cite{ebenezer2021chipqa}, video sampling was employed to acquire more efficient manual features. Liu ~\et~\cite{liu2021spatiotemporal} proposed a feature encoder for heterogeneous knowledge integration based on spatio-temporal representation learning, which directly extracts spatio-temporal features of videos, thereby mitigating the bias of individual weak labels in dataset. Li~\et~\cite{li2022blindly} utilized transfer learning to migrate from image quality assessment (IQA) to VQA, and they additionally introduced a motion feature extractor, which can be regarded as extracting temporal features.

However, limited by high computational complexity of hand-crafted feature extraction and their representation ability, the performance of these knowledge-driven BVQA methods is sub-optimal. The reasons affecting video quality are quite complicated and cannot be well captured by these handcrafted features. Therefore, the data-driven BVQA models are proposed, which consist mainly of four basic blocks, including a video pre-processor, a spatial quality analyzer, a temporal quality analyzer, and a quality regressor~\cite{sun2024analysis}. Li~\et~\cite{li2019quality} used the pre-trained ResNet50~\cite{he2016deep} to extract spatial features, then these features are fed into the GRU~\cite{cho2014learning} network for quality regression. Wu~\et~\cite{wu2023exploring} designed an objective video quality evaluator from both aesthetic and technical perspectives, where the former measures the perception of video frame distortions, while the latter assesses preferences and recommendations based on content. Chen~\et~\cite{RIRNet} incorporated the motion perception mechanism into VQA task and proposed a BVQA framework named RIRNet. Liu ~\et~\cite{liu2018end} proposed a deep learning-based multitask model called V-MEON, which utilizes a three-dimensional CNN to capture spatio-temporal information for video codec classification and quality assessment. To address the issues related to temporal distortion and the varying importance of frame content over time, Wu~\et~\cite{wu2023discovqa} proposed a spatio-temporal distortion extraction module to capture temporal distortion. Additionally, a time content transformer, resembling an encoder-decoder structure, was introduced to address concerns related to temporal attention. However, enabling end-to-end optimization of deep learning based BVQA models is not easy due to the huge demand for human-rated data and high computational cost of high resolution videos.

\subsection{Online Video Processing}
\label{subsec:obj_vqa}

Efficiency has always been one of the most important pursuits in video understanding~\cite{lin2019tsm,wang2018temporal, TRN,zolfaghari2018eco}. Lin~\et~\cite{lin2019tsm} proposed a generic temporal shift module (TSM) for efficient video understanding. Wang~\et~\cite{wang2018temporal} proposed a flexible temporal segment network (TSN), which aims to model long-range temporal structure with a segment-based sparse temporal sampling and aggregation scheme, for video action recognition. Zhou~\et~\cite{TRN} introduced an effective and interpretable network module, \ie, the temporal relation network (TRN), which is designed to learn temporal dependencies between video frames at multiple time scales. Similar to TSN, Zolfaghari~\et~\cite{zolfaghari2018eco} proposed an efficient convolutional network (ECO) with a sparse sampling strategy for video understanding.

As suggested in~\cite{FangTIP2023,sun2024analysis}, natural videos are highly redundant in both spatial and temporal dimensions, and sparsely sampled video frames are capable of obtaining competitive performance. To reduce computational cost, existing VQA approaches typically consider different sampling schemes, such as random cropping, bilinear resizing, and sptaiotemporal sampling, as the pre-processing. Wu~\et~\cite{wu2022fast} proposed a new sampling strategy for VQA that preserves both local quality and unbiased global quality with contextual relations via uniform grid mini-patch sampling (GMS). The spatially spliced and temporally aligned mini-patches named fragments are extracted as the input of video Swin Transformer~\cite{Transformer}. They later extended it to evaluate videos of any resolutions~\cite{wu2023neighbourhood}, where two important factors of spatial and temporal sampling granularity (quality-sensitive neighborhood representatives), including partitioned neighborhoods and continuous representatives, are taken into account. Similarly, Zhao~\et~\cite{zhao2023zoom} proposed a Zoom-VQA network architecture to evaluate video quality by extracting perceptual spatio-temporal features at different levels (\ie, patch level, frame level, and clip level). Liu~\et~\cite{liu2024scaling} introduced a sampling method named SAMA, which involves scaling the data into a multi-scale pyramid, conducting segment sampling at each scale, and ultimately applying masks in both temporal and spatial dimensions to adjust the size of data. In~\cite{yan2022subjective}, various sparse frame sampling strategies were employed to validate the feasibility of temporal sampling. Simultaneously, Yan~\et~\cite{yan2024revisiting} found that the spatio-temporal modeling module tends to overlook motion interruptions, indicating the viability of temporal sampling. Experimental results in~\cite{FangTIP2023} further affirmed that video perceptual quality is independent of frame order; even when disturbing the frame sequence, the performance remains robust, highlighting the operability of temporal sampling.

\section{Methodology}

\subsection{Problem Formulation}
\label{subsec:da_con}

Assume that a video is denoted by $\mathcal{V} =\{v_i\}_{i=0}^{N-1}$ with $N$ frames. An objective model $\Phi(\cdot): \mathbb{R}^{H\times W \times 3\times N}\to \mathbb{R}$, accepts $\mathcal{V}$ (or part of video frames) as input, and predicts a scalar $q$, representing the perceptual quality score of $\mathcal{V}$. Generally, the input of $\Phi(\cdot)$ includes all or most of frames in $\mathcal{V}$. As described in our previous studies~\cite{FangTIP2023, yan2024revisiting}, there exists considerable redundancy in videos, and extensive experiments showed that sparse sampling has little impact on model performance. In this paper, we take a further step to explore the effectiveness-efficiency trade-offs issue of spatio-temporal modeling in VQA models by considering spatial and temporal sampling simultaneously. The overall framework is shown in Figure~\ref{fig:motive}. We firstly squeeze $\mathcal{V}$ by sparsely temporal sampling, and can obtain a set of keyframes $\mathcal{F} = \{v_k\}_{k=0}^{K-1}$ ($k$ is the index of each keyframe), which can be formulated as:
\begin{equation}
 \mathcal{F}  = T(\mathcal{V} ; n,\eta), 
 \end{equation}
where $T (\cdot)$  denote the temporal sampling operation, and  $n$ and 
 $\eta$ represent the number of segments and the number of frames extracted from each segment, respectively. And we then apply the spatial sampling operation $S (\cdot)$ to these extracted frames, which can be formulated as:  
\begin{equation}
 \mathcal{\hat{F}}  = S(\mathcal{F}; g,s),
 \end{equation}
 where $g$ and $s$ denote the density of the grid and the size of the image patches, respectively.
Subsequently, the squeezed video $\mathcal{\hat{F}}$ is fed into $\Phi(\cdot)$ to derive the perceptual quality score $q$ as below:
\begin{equation}
 q  = \Phi(\mathcal{\hat{F}};\theta),
 \end{equation}
where $\theta$ denotes the learnable parameters of the VQA model $\Phi(\cdot)$.

\subsection{Spatio-temporal Sampling}
\label{subsec:motiv}

As aforementioned, spatio-temporal sampling in current context refers to a two-step operation for squeezing videos, including sparsely extracting frames (\ie, temporal sampling) and cropping patches from each frame (\ie, spatial sampling), and the resultant squeezed video is used as model's input. For sparsely extracting frames, we embrace the success of these related studies~\cite{wang2018temporal,lin2019tsm,zolfaghari2018eco} in video understanding, and the details will be introduced in Section~\ref{subsub:ts}. As spatial sampling, video frames' dimensions are reduced to alleviate computational burdens and enhance processing speed. This step is particularly crucial for high-resolution videos, where the original pixel number is excessively high, leading to significant consumption of computational resources. To this end, we adopt the GMS method~\cite{wu2022fast} as the default spatial sampling method. Specifically, each video frame is partitioned into uniform grids. The mini-patches with same size are randomly sampled from the uniform grids in the keyframe and then spliced together into a fragment. Given a keyframe $v \subseteq R^{3*H*W}$ (the subscript is omitted for simplicity), it is initially divided into $\beta*\beta$ grids. From each grid, the patches of size $\mu*\mu$ are selected: 
\begin{equation}
 \mathcal{M}^{\epsilon,\varepsilon}  = v[r*\epsilon:r*\epsilon+\mu, c*\varepsilon:c*\varepsilon+\mu],
 \end{equation}
where $r$ and $c$ are the starting positions of randomly selected patches in the grid, $\epsilon$ and $\varepsilon$ represent the location of the grid, respectively.
Subsequently, each patch is concatenated in the order of extraction to form a new image $\hat{v}$, with the purpose of preserving contextual relationships:
\begin{equation}
\hat{v} = \Gamma(\mathcal{M}^{\epsilon,\varepsilon}),~~0<~\epsilon,\varepsilon ~<~\beta,
 \end{equation}
where $\Gamma(\cdot)$ stands for the concatenation operation, and $\hat{v}$ denotes the frame of the squeezed video $\mathcal{\hat{F}}$.

\subsection{Spatio-temporal Modeling}
\label{subsec:stfe}

Since our main objective is to investigate online VQA model, rather than solely pursuing high-performance, we therefore resort to as simple as possible spatio-temporal feature extraction modules. Following our previous studies, we adopt the stable and effective VQA model, \ie, VSFA, in our experiments. The VSFA model uses ResNet-50 as spatial feature extractor and GRU to capture temporal quality variation, which can be described as follows:
 \begin{equation}
f_{st} = \mathcal{N}(\mathcal{\hat{F}};\theta_n),
 \end{equation}
 where $\mathcal{N}(\cdot)$ and $\theta_n$ denote the spatiotemporal feature extraction model and its pre-trained parameters, respectively. $f_{st}$ denotes the extracted spatio-temporal featrues. Note that we also implement a variant of VSFA by substituting the vanilla transformer for GRU. Moreover, the model $\mathcal{N}(\cdot)$ can be also instantiated by other light-weight modules, and the details will be introduced in Section~\ref{sec:qu_ex}.

\subsection{Global Quality Regression}

Video quality regression involves mapping the spatio-temporal features to a single scalar, we use fully connected (FC) layers to bridge those features and quality score, which is calculated as follows:
 \begin{align}
f_{st}^t &= W^{t}f_{st}^{t-1}+b^t, for~ t=1,\cdots,T,\\
q &= W^Tf_{st}^{T-1}+b^T,
 \end{align}
where $W^t$ and $b^t$ denote the $t$-th learnable parameters and bias in the global quality regression. $T$ denotes the number of FC layers, and $q$ is the predicted global quality score. 

\begin{table}[]
 \renewcommand{\arraystretch}{1.2}
\caption{The summary of the existing video quality datasets.}
\label{tab:Datasets}
\resizebox{\linewidth}{!}{
\begin{tabular}{ccccc}
\hline
Datasets  & Number  & Scenes & Type   &  Resolution \\ \hline
KoNViD-1k~\cite{hosu2017konstanz}     & 1,200      & 1,200   & In-the-wild  &  540p \\
LIVE-VQC~\cite{sinno2018large}      & 585     & 585    & In-the-wild &  240p-1080p\\
CVD2014~\cite{nuutinen2016cvd2014}       & 234        & 5      & In-the-wild  &  480p, 720p\\
LIVE-Qualcomm~\cite{ghadiyaram2017capture} & 208      & 54     & In-the-wild   &  1080p \\
LSVQ~\cite{ying2021patch}   & 38,811     & 38,811  & In-the-wild& 99p-4K    \\ 
NTIRE~\cite{liu2023ntire}   & 839     &  - & Synthetic& 720p    \\ 
\hline
\end{tabular}}
\end{table}

\begin{table*}[]
\caption{The performance of VSFA with different settings. $S_1$ indicates four 32*32 patches, $S_2$ indicates four 64*64 patches, $S_3$ indicates twenty-five 32*32 patches, and $S_4$ indicates thirty-six 32*32 patches. Its original performance on four databases is also listed and marked by $\dagger$.}
\label{tab:result1}
\resizebox{\linewidth}{!}{
\begin{tabular}{c|cc|cc|cc|cc|cc|cc}
\hline
\multirow{3}{*}{\begin{tabular}[c]{@{}c@{}}Temporal \\ and \\ Spatial\end{tabular}} & \multicolumn{6}{c}{KoNViD-1k~\cite{hosu2017konstanz} (\textbf{0.74}, \textbf{0.76})$\dagger$}  & \multicolumn{6}{|c}{CVD2014~\cite{nuutinen2016cvd2014} (\textbf{0.89}, \textbf{0.88})$\dagger$}   \\ \cline{2-13}
 & \multicolumn{2}{|c|}{TSN~\cite{wang2018temporal}} & \multicolumn{2}{c}{TSM~\cite{lin2019tsm}} & \multicolumn{2}{|c|}{ECO~\cite{zolfaghari2018eco}} & \multicolumn{2}{|c|}{TSN~\cite{wang2018temporal}} & \multicolumn{2}{c}{TSM~\cite{lin2019tsm}} & \multicolumn{2}{|c}{ECO~\cite{zolfaghari2018eco}} \\ \cline{2-13}
& PLCC  & SRCC  & PLCC  & SRCC  & PLCC  & SRCC  & PLCC  & SRCC  & PLCC  & SRCC  & PLCC  & SRCC \\ \hline
$\mathcal{S}_1$ & 0.68   & 0.69  & 0.63  & 0.65  & 0.62  & 0.64  & 0.82  & 0.84  & 0.72  & 0.73  & 0.83  & 0.83 \\
$\mathcal{S}_2$ & 0.73   & 0.72  & 0.71  & \textbf{0.71}  & 0.72  & 0.71  & 0.87  & \textbf{0.87}  & 0.73  & 0.74 & 0.85  & \textbf{0.85}\\
$\mathcal{S}_3$ & \textbf{0.74}   & \textbf{0.73}  & \textbf{0.72}  & \textbf{0.71}  & \textbf{0.74}  & \textbf{0.73}  & \textbf{0.88}  & \textbf{0.87}  & \textbf{0.80}  & \textbf{0.82}  & 0.81  & 0.79 \\
$\mathcal{S}_4$  & \textbf{0.74}   & \textbf{0.73}  & 0.70  & \textbf{0.71}  & 0.71  & 0.71  & 0.84  & 0.83  & 0.78  & 0.78  & \textbf{0.86}  & \textbf{0.85} \\ \hline
\multirow{3}{*}{\begin{tabular}[c]{@{}c@{}}Temporal \\ and \\ Spatial\end{tabular}} & \multicolumn{6}{c}{LIVE-Qualcomm~\cite{ghadiyaram2017capture} (\textbf{0.73}, \textbf{0.74})$\dagger$}  & \multicolumn{6}{|c}{LIVE-VQC~\cite{sinno2018large} (\textbf{0.77}, \textbf{0.72})$\dagger$}   \\ \cline{2-13}
 & \multicolumn{2}{|c|}{TSN~\cite{wang2018temporal}} & \multicolumn{2}{c}{TSM~\cite{lin2019tsm}} & \multicolumn{2}{|c|}{ECO~\cite{zolfaghari2018eco}} & \multicolumn{2}{|c|}{TSN~\cite{wang2018temporal}} & \multicolumn{2}{c}{TSM~\cite{lin2019tsm}} & \multicolumn{2}{|c}{ECO~\cite{zolfaghari2018eco}} \\ \cline{2-13}
 & PLCC  & SRCC  & PLCC  & SRCC  & PLCC  & SRCC  & PLCC  & SRCC  & PLCC  & SRCC  & PLCC  & SRCC \\ \hline
$\mathcal{S}_1$ & 0.72  & 0.71  & 0.57   & 0.52  & 0.66  & 0.65 & 0.72 & 0.68 & 0.64 & 0.63  & \textbf{0.75}  & \textbf{0.73} \\ 
$\mathcal{S}_2$& 0.70  & 0.69  & 0.54   & 0.54  & 0.71  & 0.68 &\textbf{0.78} & 0.72 & 0.73 & 0.68  & 0.71  & 0.66 \\ 
$\mathcal{S}_3$ & 0.76  & 0.73  & 0.55   & 0.57 & \textbf{0.77}  & \textbf{0.78} & 0.75 & \textbf{0.73} & \textbf{0.74} & \textbf{0.71}  & \textbf{0.75}  & 0.71 \\ 
$\mathcal{S}_4$& \textbf{0.82}  & \textbf{0.81}  & \textbf{0.68}   & \textbf{0.72}  & 0.73  & 0.74  & 0.75 & \textbf{0.73} & 0.72 & 0.69  & \textbf{0.75}  & 0.72  \\ \hline
\multirow{3}{*}{\begin{tabular}[c]{@{}c@{}}Temporal \\ and \\ Spatial\end{tabular}} & \multicolumn{6}{c}{LSVQ~\cite{ying2021patch}}  & \multicolumn{6}{|c}{LSVQ$_{1080P}$~\cite{ying2021patch}}   \\ \cline{2-13}
 & \multicolumn{2}{|c|}{TSN~\cite{wang2018temporal}} & \multicolumn{2}{c}{TSM~\cite{lin2019tsm}} & \multicolumn{2}{|c|}{ECO~\cite{zolfaghari2018eco}} & \multicolumn{2}{|c|}{TSN~\cite{wang2018temporal}} & \multicolumn{2}{c}{TSM~\cite{lin2019tsm}} & \multicolumn{2}{|c}{ECO~\cite{zolfaghari2018eco}} \\ \cline{2-13}
 & PLCC  & SRCC  & PLCC  & SRCC  & PLCC  & SRCC  & PLCC  & SRCC  & PLCC  & SRCC  & PLCC  & SRCC \\ \hline
 $\mathcal{S}_1$ & 0.71   & 0.71  & 0.62  & 0.62  & 0.65  & 0.65  & 0.61  & 0.59  & 0.55  & 0.54  & 0.58  & 0.55 \\ 
 $\mathcal{S}_2$ & 0.76   & 0.76  & 0.71  & 0.71  & 0.71  & 0.71  & \textbf{0.67}  & \textbf{0.64}  & 0.61  & 0.58  & 0.63  & 0.60 \\ 
 $\mathcal{S}_3$& \textbf{0.77}   & 0.77  & 0.74  & 0.74  & 0.74  & 0.75  & 0.65  & 0.62  & 0.62  & 0.59  & 0.63  & 0.60  \\
 $\mathcal{S}_4$&\textbf{0.77}   &\textbf{0.78}  & \textbf{0.75}  & \textbf{0.76}  & \textbf{0.76}  & \textbf{0.76}  & 0.66  & \textbf{0.64}  & \textbf{0.63}  & \textbf{0.60}  & \textbf{0.64}  & \textbf{0.63}  \\ \hline
\end{tabular}}
\end{table*}

\begin{table*}[]
\caption{The performance of TransformerVSFA. $S_1$ indicates four 32*32 patches, $S_2$  indicates four 64*64 patches, $S_3$ indicates twenty-five 32*32 patches, and $S_4$ indicates thirty-six 32*32 patches. Its original performance on four databases is also listed and marked by $\dagger$.}
\label{tab:result2}
\resizebox{\linewidth}{!}{
\begin{tabular}{c|cc|cc|cc|cc|cc|cc}
\hline
\multirow{3}{*}{\begin{tabular}[c]{@{}c@{}}Temporal \\ and \\ Spatial\end{tabular}} & \multicolumn{6}{c}{KoNViD-1k~\cite{hosu2017konstanz} (\textbf{0.79}, \textbf{0.78})$\dagger$}  & \multicolumn{6}{|c}{CVD2014~\cite{nuutinen2016cvd2014} (\textbf{0.87}, \textbf{0.86})$\dagger$}   \\ \cline{2-13}
 & \multicolumn{2}{|c|}{TSN~\cite{wang2018temporal}} & \multicolumn{2}{c}{TSM~\cite{lin2019tsm}} & \multicolumn{2}{|c|}{ECO~\cite{zolfaghari2018eco}} & \multicolumn{2}{|c|}{TSN~\cite{wang2018temporal}} & \multicolumn{2}{c}{TSM~\cite{lin2019tsm}} & \multicolumn{2}{|c}{ECO~\cite{zolfaghari2018eco}} \\ \cline{2-13}
& PLCC  & SRCC  & PLCC  & SRCC  & PLCC  & SRCC  & PLCC  & SRCC  & PLCC  & SRCC  & PLCC  & SRCC \\ \hline
$\mathcal{S}_1$ & 0.70  & 0.70 & 0.64  & 0.65  & 0.67  & 0.67  & 0.82  & 0.82  & 0.70  & 0.70  & 0.80  & 0.78 \\
$\mathcal{S}_2$ & 0.74 & 0.74  & 0.71  &0.71  & 0.70  & 0.69  & \textbf{0.88}  & \textbf{0.88}  & 0.76  & 0.75  & 0.85  & \textbf{0.86}\\
$\mathcal{S}_3$ &  0.74 &0.74 & 0.71  & 0.70  & \textbf{0.72}  & 0.71  & 0.87  & \textbf{0.88}  & \textbf{0.84}  & \textbf{0.83}  & \textbf{0.86}  & \textbf{0.86}\\
$\mathcal{S}_4$   & \textbf{0.75} & \textbf{0.75}  & \textbf{0.73}  & \textbf{0.73}  & \textbf{0.72}  & 0.72  & \textbf{0.88}  & \textbf{0.88}  & 0.83  & \textbf{0.83}  & 0.84  & 0.83 \\ \hline
\multirow{3}{*}{\begin{tabular}[c]{@{}c@{}}Temporal \\ and \\ Spatial\end{tabular}} & \multicolumn{6}{c}{LIVE-Qualcomm~\cite{ghadiyaram2017capture} (\textbf{0.69}, \textbf{0.63})$\dagger$}  & \multicolumn{6}{|c}{LIVE-VQC~\cite{sinno2018large} (\textbf{0.72}, \textbf{0.66})$\dagger$}   \\ \cline{2-13}
 & \multicolumn{2}{|c|}{TSN~\cite{wang2018temporal}} & \multicolumn{2}{c}{TSM~\cite{lin2019tsm}} & \multicolumn{2}{|c|}{ECO~\cite{zolfaghari2018eco}} & \multicolumn{2}{|c|}{TSN~\cite{wang2018temporal}} & \multicolumn{2}{c}{TSM~\cite{lin2019tsm}} & \multicolumn{2}{|c}{ECO~\cite{zolfaghari2018eco}} \\ \cline{2-13}
 & PLCC  & SRCC  & PLCC  & SRCC  & PLCC  & SRCC  & PLCC  & SRCC  & PLCC  & SRCC  & PLCC  & SRCC \\ \hline
$\mathcal{S}_1$  & 0.72  & 0.72  & \textbf{0.72}  & 0.44  & 0.68  & 0.68 & 0.71 & 0.69  & 0.70  & 0.67  & 0.68  & 0.65\\ 
$\mathcal{S}_2$ & 0.69  & 0.63  & 0.65  & 0.63  & 0.68  & 0.61& 0.74 & 0.72  & 0.74  & 0.72  & 0.71  & 0.69\\ 
$\mathcal{S}_3$  & \textbf{0.74}  & \textbf{0.75}  & 0.64  & \textbf{0.68}  & \textbf{0.75}  & \textbf{0.72} & \textbf{0.77} & \textbf{0.73}  & \textbf{0.77}  & \textbf{0.73}  & \textbf{0.73}  & \textbf{0.71} \\ 
$\mathcal{S}_4$ & 0.71  & 0.71  & 0.70  & \textbf{0.68}  & 0.72  & 0.71  & 0.75 & 0.72  & 0.75  & 0.72  & 0.72  & 0.69 \\ \hline
\multirow{3}{*}{\begin{tabular}[c]{@{}c@{}}Temporal \\ and \\ Spatial\end{tabular}} & \multicolumn{6}{c}{LSVQ~\cite{ying2021patch}}  & \multicolumn{6}{|c}{LSVQ$_{1080P}$~\cite{ying2021patch}}   \\ \cline{2-13}
 & \multicolumn{2}{|c|}{TSN~\cite{wang2018temporal}} & \multicolumn{2}{c}{TSM~\cite{lin2019tsm}} & \multicolumn{2}{|c|}{ECO~\cite{zolfaghari2018eco}} & \multicolumn{2}{|c|}{TSN~\cite{wang2018temporal}} & \multicolumn{2}{c}{TSM~\cite{lin2019tsm}} & \multicolumn{2}{|c}{ECO~\cite{zolfaghari2018eco}} \\ \cline{2-13}
 & PLCC  & SRCC  & PLCC  & SRCC  & PLCC  & SRCC  & PLCC  & SRCC  & PLCC  & SRCC  & PLCC  & SRCC \\ \hline
 $\mathcal{S}_1$ & 0.69  & 0.69  & 0.63  & 0.63  & 0.65  & 0.66  & 0.59  & 0.57  & 0.54  & 0.52  & 0.57  & 0.54  \\ 
 $\mathcal{S}_2$   & 0.75  & 0.75  & 0.70  & 0.70  & 0.71  & 0.72  & \textbf{0.64}  & \textbf{0.61}  & 0.59  & 0.56  & 0.61  & 0.58 \\ 
 $\mathcal{S}_3$  & 0.76  & 0.76  & 0.74  & 0.74  & 0.74  & 0.75  & \textbf{0.64}  & 0.60  & 0.61  & 0.57  & 0.62  & 0.59 \\
 $\mathcal{S}_4$ & \textbf{0.77}  & \textbf{0.77}  & \textbf{0.76}  & \textbf{0.76}  & \textbf{0.76}  & \textbf{0.76}  & \textbf{0.64}  & \textbf{0.61}  & \textbf{0.63}  & \textbf{0.60}  & \textbf{0.63}  & \textbf{0.60} \\ \hline
\end{tabular}}
\end{table*}

\begin{table}[]
 \renewcommand{\arraystretch}{1.2}
\caption{
The performance of VSFA and TransformerVSFA on the NTIRE dataset. $\mathcal{S}_1$ denotes four 32*32 patches, $\mathcal{S}_2$ denotes four 64*64 patches, $\mathcal{S}_3$ denotes twenty-five 32*32 patches, and $\mathcal{S}_4$ denotes thirty-six 32*32 patches.}
\label{tab:result3}
\begin{tabular}{c|ll|ll|ll}
\hline
\multirow{3}{*}{\begin{tabular}[c]{@{}c@{}}Temporal \\ and \\ Spatial\end{tabular}} & \multicolumn{6}{c}{VSFA~\cite{li2019quality}}\\ \cline{2-7} & \multicolumn{2}{c|}{TSN~\cite{wang2018temporal}}& \multicolumn{2}{c|}{TSM~\cite{lin2019tsm}}& \multicolumn{2}{c}{ECO~\cite{zolfaghari2018eco}}\\ \cline{2-7} 
& \multicolumn{1}{c}{PLCC} & \multicolumn{1}{c|}{SRCC} & \multicolumn{1}{c}{PLCC} & \multicolumn{1}{c|}{SRCC} & \multicolumn{1}{c}{PLCC} & \multicolumn{1}{c}{SRCC} \\ \hline
$\mathcal{S}_1$& 0.76 & 0.77 & 0.60 & 0.61 & 0.62 & 0.64                      \\ 
$\mathcal{S}_2$& 0.77 &0.78 & 0.66 & 0.67 & \textbf{0.73} & \textbf{0.74}                      \\ 
$\mathcal{S}_3$& 0.69 & 0.72 & 0.68 & 0.70 & 0.72 & 0.73                      \\ \
$\mathcal{S}_4$& \textbf{0.73} & \textbf{0.74} & \textbf{0.70} & \textbf{0.72} & 0.71 & 0.73                      \\ \hline
\multirow{3}{*}{\begin{tabular}[c]{@{}c@{}}Temporal \\ and \\ Spatial\end{tabular}} & \multicolumn{6}{c}{TransformerVSFA~\cite{vaswani2017attention}}\\ \cline{2-7} & \multicolumn{2}{c|}{TSN~\cite{wang2018temporal}}& \multicolumn{2}{c|}{TSM~\cite{lin2019tsm}}& \multicolumn{2}{c}{ECO~\cite{zolfaghari2018eco}}\\ \cline{2-7} 
& \multicolumn{1}{c}{PLCC} & \multicolumn{1}{c|}{SRCC} & \multicolumn{1}{c}{PLCC} & \multicolumn{1}{c|}{SRCC} & \multicolumn{1}{c}{PLCC} & \multicolumn{1}{c}{SRCC} \\ \hline
$\mathcal{S}_1$& 0.72 & 0.73 & 0.61 & 0.62 & 0.62 & 0.66                      \\ 
$\mathcal{S}_2$& \textbf{0.77} & \textbf{0.78} & 0.68 & 0.69 & \textbf{0.75} & \textbf{0.77}                      \\ 
$\mathcal{S}_3$& 0.73 & 0.76 & 0.70 & 0.69 & 0.70 & 0.71                      \\ 
$\mathcal{S}_4$& 0.76 & 0.78 & \textbf{0.75} &\textbf{0.75} & 0.71 & 0.72                      \\ \hline
\end{tabular}
\end{table}

\section{Quantitative Experiments}
\label{sec:qu_ex}

\subsection{Experimental Settings}
\label{subsec:ex_se}

\subsubsection{Test Databases}
We evaluate BVQA models on six VQA datasets: KoNViD-1k~\cite{hosu2017konstanz}, LIVE-VQC~\cite{sinno2018large}, CVD2014~\cite{nuutinen2016cvd2014}, LIVE-Qualcomm~\cite{ghadiyaram2017capture}, LSVQ~\cite{ying2021patch}, and NTIRE~\cite{liu2023ntire}. Table~\ref{tab:Datasets} summarizes the key details of these test datasets. For each dataset, we randomly split the data into 60\% for training, 20\% for validation, and 20\% for testing. To ensure fair comparisons, we perform three rounds of training-validation-testing splits, and report the average test performance. It should be noted that we evaluate the model's performance on the LSVQ$_{1080P}$ dataset using the corresponding parameters trained on the LSVQ dataset.

\subsubsection{Training and Test Details}
We select a representative and stable BVQA model, \ie, VSFA~\cite{li2019quality}, as the backbone, and also test its variant by substituting its GRU module with the vanilla transformer~\cite{vaswani2017attention}, which is denoted by TransformerVSFA in this paper. For feature extraction, the ResNet50 pre-trained on the ImageNet dataset~\cite{russakovsky2015imagenet} is adopted as feature extractor. The other settings are consistent with~\cite{li2019quality}. The size of the spatially sampled image patches is set to 32*32 and 64*64, and the number of sampled patches is set to 4, 25, and 36. With the combination of different patch sizes and numbers, we get four spatial sampling schemes: (\uppercase\expandafter{\romannumeral1}) $\mathcal{S}_1$: the patch size is 32*32 and number is 4; (\uppercase\expandafter{\romannumeral2}) $\mathcal{S}_2$: the patch size is 64*64 and number is 4; (\uppercase\expandafter{\romannumeral3}) $\mathcal{S}_3$: the patch size is 32*32 and number is 25; (\uppercase\expandafter{\romannumeral4}) $\mathcal{S}_4$: the patch size is 32*32 and number is 36. Additionally, we further design a light weight VQA model based on the design principle of VSFA, where we adopt the proposed spatio-temporal sampling method for extracting spatio-temporal patches, a fast spatial distortion capturer, a compact spatio-temporal distortion perception module, and a global quality regression module, and other details will be introduced in Section~\ref{subsec:as}.

In order to measure the performance of the test BVQA models, two metrics, including Pearson linear correlation coefficient (PLCC) and Spearman's rank-order correlation coefficient (SRCC), are used. PLCC evaluates the prediction accuracy of the model, while SRCC computes the prediction monotonicity. As suggested in~\cite{video2005final}, we use a four-parameter logistic function to map the predicted objective score to the subjective score before calculating the PLCC:

 \begin{equation}f(o) = \frac{\tau_1 - \tau_2}{1+e^{-\frac{o-\tau_3}{\tau_4}\quad}}+\tau_2,
 \end{equation}
where $\{\tau_{1}, \tau_{2}, \tau_{3}, \tau_{4}\}$ are the parameters to be fitted.

\subsubsection{Temporal Sampling}
\label{subsub:ts}

\begin{figure}
\centering

\subfigure[TSN]{
\begin{minipage}[]{\linewidth}
\includegraphics[width=1\linewidth]{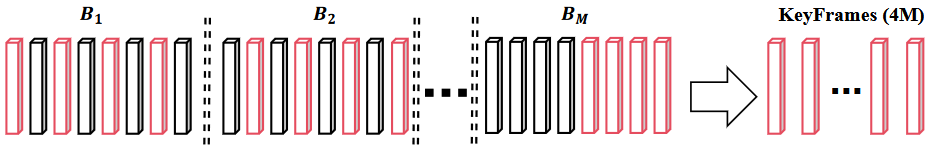}\vspace{0.5pt}
\end{minipage}
}
\subfigure[TSM]{
\begin{minipage}[]{\linewidth}
\includegraphics[width=1\linewidth]{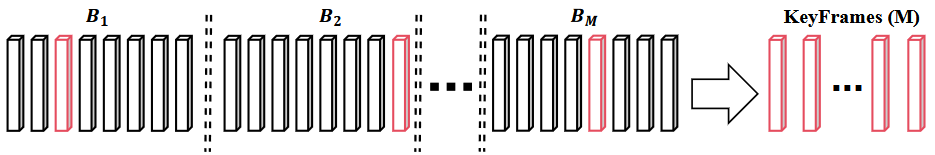}\vspace{0.51pt}
\end{minipage}
}
\subfigure[ECO]{
\begin{minipage}[]{\linewidth}
\includegraphics[width=1\linewidth]{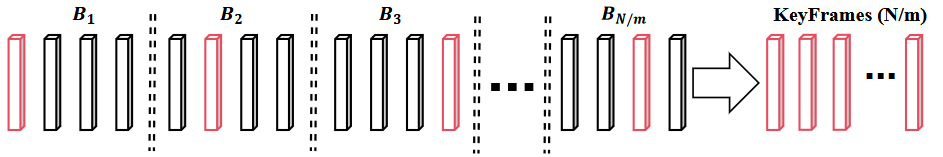}\vspace{0.51pt}
\end{minipage}
}
\caption{The intuitive comparison of different temporal sampling methods. Both TSN and TSM divide a video into $M$ segments, and their difference is that TSM extracts four frames (default setting in this paper) at a fixed length for each segment while TSM only samples one frame from each segment. The difference between TSM and ECO is that TSM has a fixed number of segments that is determined by the number of video frames of each video, while ECO has a varying number of segments for different videos. ($M$represents the number of video blocks, $N$ denotes the total number of video frames, and $m$ indicates the number of frames contained in each block.)}
\label{fig:tem}
\end{figure}

We test three classical temporal sampling methods, including TSN~\cite{wang2018temporal}, TSM~\cite{lin2019tsm}, and ECO~\cite{zolfaghari2018eco}. The intuitive comparison of these sampling methods are depicted in Figure~\ref{fig:tem}, and their details are described as below.

\textit{\textbf{TSN}}. It has been widely used to effectively extract keyframes from a given video $\mathcal{V} =\{v_i\}_{i=0}^{N-1}$. By setting a hyperparameter $\eta$, TSN divides a video uniformly into $M$ segments, each segment containing approximately $\theta = \lfloor N/M \rfloor$ video frames, where $\lfloor \cdot \rfloor$ denotes the floor function, and $M$ and $\eta$ are set to 10 and 4, respectively. To further extract keyframes from each segment, we set the sampling step $step$ to $\lfloor (\theta-1)/(\eta-1) \rfloor$, ensuring the even distribution of keyframes within each segment:
\begin{equation}
	\Omega = \begin{cases}
	{\rm Range}(1, {\rm min} (2+step * (\eta-1),\\
        ~~~~~~~~~~~~\theta+1), step), &\text{if}~step>0, \\
	[1] * \eta, &\text{if}~step < = 0	,
		   \end{cases}
\end{equation}
where $\rm Range(\cdot)$ is a function that generates a list based on the start value, stop value, and step size. This method can flexibly adapt to videos of different lengths by properly selecting segment divisions and step sizes, effectively extracting keyframes. Finally, all the keyframes extracted from the original video are
\begin{equation}
\mathcal{T} = \{\mathcal{T}^0, \mathcal{T}^2,\cdots, \mathcal{T}^{M-1}\},
\end{equation}
\begin{equation}
\mathcal{T}^m = \{v_{\Omega[0]+\theta*m}, v_{\Omega[1]+\theta*m},\cdots,v_{\Omega[\eta-1]+\theta*m}\},
\end{equation}
where $m$ is the index of segments and $m\in\{0,1,\cdots,M-1\}$, $\Omega[\cdot]$ denotes the element in set $\Omega$. This TSN-based keyframe extraction method ensures both the representativeness of keyframes and processing efficiency, which is crucial for subsequent video analysis and processing tasks.

\textit{\textbf{TSM}}.
The core idea of TSM is to promote the interaction of information between adjacent frames by moving a portion of channels along the temporal dimension. Assuming the original video is divided into $M$ segments based on the parameter $\xi$, where $M$ is set to 10. The indices of the sampled keyframes can be denoted as: 
\begin{equation}
\mathcal{T} =\{v_{m*\xi+\epsilon_m}\}_{m=0}^{M-1},
\end{equation}
where $\epsilon_m$ represents a random value within the range $[0, \xi)$ for the $m$-th segment. Typically, $\xi$ is set to $\lfloor N/M \rfloor$ to ensure the video is evenly segmented into $M$ parts, and each segment covers a sufficient range of frames to ensure the sampled keyframes are representative. The TSM-based keyframe sampling method effectively enhances the efficiency of information exchange in the video, providing a richer information foundation for subsequent quality prediction.

\textit{\textbf{ECO}}. 
When averaging the number of video frames by obtained segments, ECO behaves similarly to TSM. However, the number of fragments $M$ is set to 20. That is, ECO randomly selects a keyframe from each segment based on the parameter $\zeta$ as:
\begin{equation}
	\mathcal{T} = \begin{cases}
	\{v_{m*\zeta+\epsilon_m}\}_{m=0}^{M-1}, &\text{if}~(N\bmod M)\le0 \\
	\{v_{m*\zeta+\epsilon_m}\}_{m=0}^{M}, &\text{if}~(N\bmod M)\neq0	
		   \end{cases}.
\end{equation}
In contrast, another keyframe is extracted from the last unevenly divided keyframe. Sampling keyframes in this way ensures an even distribution throughout the entire video sequence, thereby ensuring comprehensive coverage of the video content and accurate representation.

\subsection{Experimental Results and Analysis}

The experimental results are presented in Tables \ref{tab:result1}, \ref{tab:result2}, and \ref{tab:result3}. As can be seen from these tables, the proposed spatio-temporal sampling does not lead to a significant drop in performance across the datasets, demonstrating the viability of online VQA with reduced video frame input. While TSN achieves the highest performance, it requires extracting more video frames and leads to higher computational cost. In contrast, TSM samples significantly fewer frames while still delivering robust results. On the KoNViD-1k, LIVE-VQC, and LSVQ datasets, despite TSM's reduced sampling, VSFA's performance meets or exceeds expectations. This could be attributed to the simpler video content in these datasets~\cite{sun2024analysis}, where the temporal features are less critical to predict overall video quality. Consequently, the model’s performance could be further enhanced by adopting more sophisticated data sampling strategies. Besides, the model performs better on the CVD2014 dataset. The videos in CVD2014 are recorded with nearly stationary cameras, resulting in minimal temporal distortion. In contrast, NTIRE consists of diverse algorithm-related distortions, which introduces a new challenge for quality prediction, especially when temporal continuity is largely disrupted (\ie, use the TSM as a temporal sampling method). The LSVQ$_{1080p}$ dataset, consisting of high-resolution videos with relatively high quality, demonstrates a notable impact of spatial sampling on performance, indicating the presence of considerable redundant information within the video content. Reducing this redundancy can increase model's efficiency. Doubling the spatial sampling range leads to a significant performance improvement, suggesting that expanding the sampling range can enhance the model's accuracy. However, with high-resolution videos, there is no significant difference between two sampling methods, possibly because both sampling methods have a relatively smaller impact compared to the overall width and height of the video. 

In summary, VSFA with aggressive spatiotemporal sampling shows a small performance decline in the CVD2014, KoNViD-1k and LSVQ databases, which is attributed to the fact that these VQA databases are dominated by spatial distortion, and the strong spatial feature extraction, \ie, ResNet-50, is sufficient to handle these databases. In contrast, VSFA with the aggressive sampling operation presents a significant performance drop on the temporal distortion dominated databases, \ie, LIVE-Qualcomm, and LIVE-VQC. Note that the VSFA with the TSN and $\mathcal{S}_4$ sampling scheme obtains a large gain on the LIVE-Qualcomm database, whose possible reason is that inputting more patches contributes to capture the temporal distortion caused by spatial distortion. Compared with the results on these aforementioned databases, the VSFA behaves far from satisfactory on the LSVQ$_{1080P}$ and NTIRE databases, since the former contains high-resolution videos with relatively high quality, making the objective model hard to distinguish between them, and the latter contains the complex and challenging algorithm-related distortions. The experimental results for spatio-temporal sampling provides a deeper understanding of how VQA models perform under various sampling strategies, and offers valuable insights for designing more efficient and accurate VQA models. In practical applications, appropriate data sampling strategies can be selected based on specific scenarios and requirements to balance performance and computational resources. The experimental results from both temporal and spatial sampling offer valuable insights for optimizing VQA models, shedding light on designing online VQA models. Furthermore, it is should be noted that both temporal and spatial sampling might miss critical information in handling high-motion or highly dynamic video content, since these two types of videos show different characteristics from that in the test databases (also being widely used in VQA studies~\cite{zhu2020faster,FangTIP2023,sun2024analysis}) in this paper. Specifically, the CVD2014, KoNViD-1k, and LSVQ databases are dominated by spatial distortion, the LIVE-Qualcomm and LIVE-VQC databases suffer from temporal distortion, the LSVQ$_{1080P}$ contains high-resolution videos with relatively high quality (which is difficult to distinguish), and the NTIRE database contains complex and challenging algorithm-related distortions. However, for those high-motion databases~\cite{chen2019qoe,shang2021study}, which involve rapidly moving objects, frequent scene changes, complex actions, or quickly shifting visuals, directly applying the proposed model might be no longer suitable due to the loss of temporal information and the weak temporal modeling component and simple modeling strategy~\cite{sun2024analysis}, and more elaborate designs, \eg, large-scale pre-training or awesome modules for capturing motion features, should be required. For those highly dynamic videos~\cite{shang2023study, saini2024hidro}, the visible distortions in the dark and bright zones and the concomitant temporal distortion bring a big challenge to the effectiveness of spatial and temporal sampling. Therefore, the generalization and application of the proposed model is an open problem that deserves more effort to investigate.

\subsection{A Further Design of Online VQA Model}
\label{subsec:as}

Considering that the VSFA and TransformerVSFA models still rely on ResNet50, we make a further step to design an online VQA model by integrating \textbf{M}obileNet~\cite{howard2017mobilenets} and \textbf{G}raph network~\cite{sun2022graphiqa}, and the proposed online VQA model is named MGQA. The experimental results of the MGQA model (the default spatial and temporal sampling methods are $\mathcal{S}_3$ and ECO, respectively) on various datasets are shown in Table~\ref{tab:result7}. The model consistently performs well, demonstrating state-of-the-art results on the CVD2014 and LIVE-Qualcomm datasets, on par with other lightweight models. MobileNet, which is designed for mobile and embedded visual applications, is noteworthy for its high efficiency, reducing computational complexity through depth-wise separable convolution while maintaining strong feature extraction capabilities. This makes it particularly effective at extracting spatial features from videos as input to the quality regression module.

\begin{figure}[htbp]
	\centering
	\subfigure[Parameters] {\includegraphics[width=0.42\textwidth]{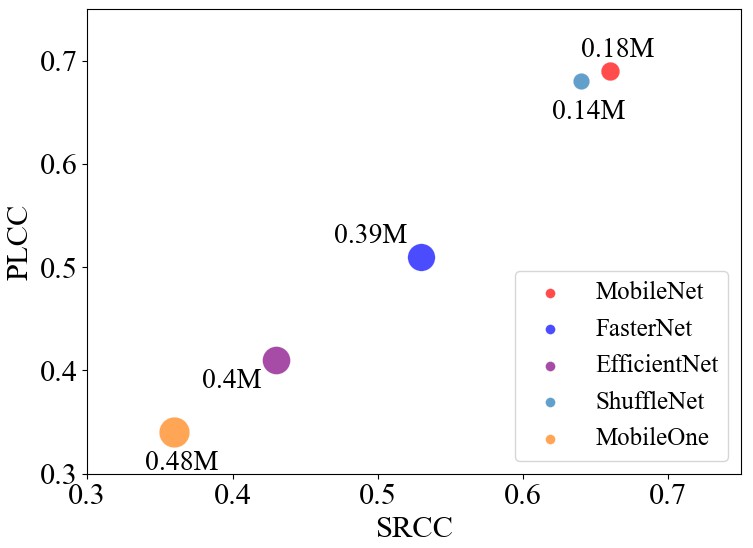}}
	\subfigure[Flops] {\includegraphics[width=0.42\textwidth]{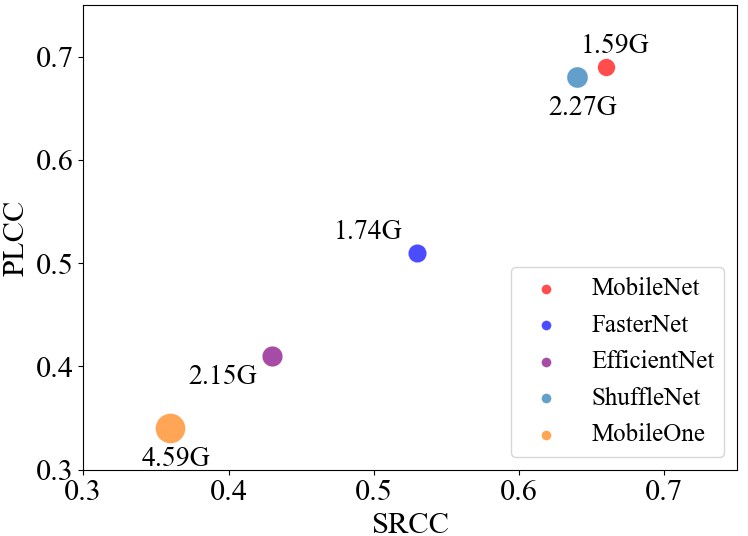}}
	\caption{Comparison of the proposed MGQA with different spatial feature extractors in terms of parameters and Flops.}
	\label{fig:pf}
\end{figure}

Additionally, different lightweight spatial feature extractors and quality regressors are used to extract spatial features with sampling schemes $\mathcal{S}_3$ and ECO. The experimental results, shown in Tables \ref{tab:result8},~\ref{tab:result8_1}, and Figure~\ref{fig:pf}, suggest that among various lightweight models, MobileNet significantly reduces parameters and Flops compared to other similar models. The MobileNet feature extractor also maintains strong performance even after preprocessing, likely due to the width multiplier and resolution multiplier hyperparameters. These hyperparameters allow MobileNet to efficiently extract fine-grained features, making it effective for computational visual tasks. MobileNet's advantage likely stems from its unique structural design, particularly in how it uses width and resolution multipliers. These settings enable MobileNet to capture key image features while keeping the computational cost low. This design strategy is crucial for visual tasks, as it supports strong performance while preserving a lightweight model structure. For the quality regressor, the Grap and fully connected layers (FCs) can obtain better performance. And we also test the sensitivity of MGQA to spatio-temporal sampling strategies, and the experimental results are shown in Table~\ref{tab:result8_2}. From Table~\ref{tab:result8_2}, we can clearly observe that MGQA suffers from a large performance decrease when the input data is reduced sharply, and performs stably when using the $\mathcal{S}_3$ spatial sampling method.

\begin{table}[]
  \renewcommand{\arraystretch}{1.2}
\centering
\caption{The performance of the proposed MGQA model on the test datasets using spatial sampling scheme $\mathcal{S}_3$ and temporal sampling ECO.}
\label{tab:result7}
 \setlength{\tabcolsep}{5mm}{
\begin{tabular}{c|cc} \hline
& \multicolumn{2}{c}{MGQA} \\ \cline{2-3}
\multirow{-2}{*}{Datasets}& PLCC                & SRCC                \\ \hline
KoNViD-1k~\cite{hosu2017konstanz}& 0.74                & 0.74                \\
LIVE-VQC~\cite{sinno2018large} & 0.69                & 0.66               \\
CVD2014~\cite{nuutinen2016cvd2014}  & 0.87                & 0.89                \\
LIVE-Qualcomm~\cite{ghadiyaram2017capture}& 0.83                & 0.86                \\
NTIRE~\cite{liu2023ntire}& 0.75                & 0.76                \\ 
 \hline
\end{tabular}}
\end{table}

\begin{table}[]
\centering
\caption{The results of the proposed MGQA with different lightweight spatial feature extraction models on the LIVE-VQC database.}
\label{tab:result8}
 \setlength{\tabcolsep}{5mm}{
\begin{tabular}{c|cc}
\hline
\multirow{2}{*}{Backbone}  & \multicolumn{2}{c}{LIVE-VQC~\cite{sinno2018large}}    \\ \cline{2-3} 
& \multicolumn{1}{c|}{PLCC} & SRCC \\ \hline
MobileNet~\cite{howard2017mobilenets}                     & \multicolumn{1}{c|}{\textbf{0.69}} & \textbf{0.66} \\ 
FasterNet~\cite{chen2023run}                                                  & \multicolumn{1}{c|}{0.51} & 0.53 \\ 
EfficientNet~\cite{tan2019efficientnet}                                                & \multicolumn{1}{c|}{0.41} & 0.43 \\  
ShuffleNet~\cite{zhang2018shufflenet}                                                  & \multicolumn{1}{c|}{0.68} & 0.64 \\  
MobileOne~\cite{vasu2023mobileone}                                                 & \multicolumn{1}{c|}{0.34} & 0.36 \\ \hline
\end{tabular}}
\end{table}

\begin{table}[]
\centering
\caption{The results of the proposed MGQA with different regressors on the LIVE-VQC database.}
\label{tab:result8_1}
 \setlength{\tabcolsep}{4mm}{
\begin{tabular}{c|cc}
\hline
\multirow{2}{*}{ Regressor} & \multicolumn{2}{c}{LIVE-VQC~\cite{sinno2018large}}    \\ \cline{2-3} 
& \multicolumn{1}{c|}{PLCC} & SRCC \\ \hline
  Graph~\cite{sun2022graphiqa}          & \multicolumn{1}{c|}{\textbf{0.69}} & \textbf{0.66} \\ 
                 GRU~\cite{cho2014learning}                               & \multicolumn{1}{c|}{0.27} & 0.16 \\ 
                 LSTM~\cite{graves2012long}                              & \multicolumn{1}{c|}{0.21} & 0.03 \\  
                 Transofrmer~\cite{vaswani2017attention}                        & \multicolumn{1}{c|}{0.47} & 0.44 \\  
                FC (Full-Connection)                        & \multicolumn{1}{c|}{0.65} & 0.64 \\  
 \hline
\end{tabular}}
\end{table}

\begin{table}[]
\centering
\caption{The Performance of the proposed MGQA with different settings. ($S_1$ indicates four 32*32 patches, $S_2$ indicates four 64*64 patches, and  $S_3$ indicates twenty-five 32*32 patches.)}
\label{tab:result8_2}
\setlength{\tabcolsep}{6.5mm}{
\begin{tabular}{c|c|cc}
\hline
\multirow{2}{*}{Spatial}   & \multirow{2}{*}{Temporal} & \multicolumn{2}{c}{LIVE-VQC~\cite{sinno2018large}}    \\ \cline{3-4} 
                     &                    & \multicolumn{1}{c}{PLCC} & SRCC \\ \hline
                     \multirow{3}{*}{$S_1$} & TSN                & \multicolumn{1}{c}{0.54} & 0.57 \\ 
                     & TSM                & \multicolumn{1}{c}{0.48} & 0.47 \\ 
                     & ECO                & \multicolumn{1}{c}{0.47} & 0.53 \\ \hline

\multirow{3}{*}{$S_2$} & TSN                & \multicolumn{1}{c}{0.65} & 0.62 \\ 
                     & TSM                & \multicolumn{1}{c}{0.62} & 0.64 \\ 
                     & ECO                & \multicolumn{1}{c}{0.60} & 0.63 \\ \hline
\multirow{3}{*}{$S_3$} & TSN                & \multicolumn{1}{c}{0.68} & 0.64 \\ 
                     & TSM                & \multicolumn{1}{c}{0.67} & 0.64 \\ 
                     & ECO                & \multicolumn{1}{c}{\textbf{0.69}} & \textbf{0.66} \\ \hline
\end{tabular}}
\end{table}

\begin{table}[]
\centering
\caption{The comparison of computational cost on six video quality databases between the VSFA and MGQA models in terms of processed data, and the input sizes of these two models are also listed.}
\label{tab:result10}
\setlength{\tabcolsep}{2mm}{
\begin{tabular}{c|c|c|c} \hline
Datasets & KoNViD-1k~\cite{hosu2017konstanz}  & CVD2014~\cite{nuutinen2016cvd2014} & LIVE-Qualcomm~\cite{ghadiyaram2017capture} \\ \hline
VSFA     & 208*540*960  & 337*540*960  & 450*1080*1920 \\  
MGQA     & 10*160*160   & 16*160*160   & 22*160*160    \\
Ratio    & \textbf{99.76\%} $\downarrow$      & \textbf{99.77\%} $\downarrow$     & \textbf{99.93}\% $\downarrow$      \\ \hline
Datasets & LIVE-VQC~\cite{sinno2018large}     & LSVQ~\cite{ying2021patch}         & NTIRE~\cite{liu2023ntire}         \\ \hline
VSFA     & 300*720*1280 & 240*720*1280 & 270*720*1280  \\
MGQA     & 15*160*160   & 12*160*160   & 14*160*160    \\
Ratio    & \textbf{99.86\%}$\downarrow$   & \textbf{99.86\%}  $\downarrow$    & \textbf{99.85\%}$\downarrow$   \\
\hline
\end{tabular}}
\end{table}

\subsection{Ablation Study}
\label{subsec:as}

\begin{table}[]
 \renewcommand{\arraystretch}{1.2}
\centering
\caption{
The performance of the MGQA model on the KoNViD-1k and LIVE-VQC datasets using $\mathcal{S}_3$ scheme.}
\label{tab:result9}
 \setlength{\tabcolsep}{2.5mm}{
\begin{tabular}{c|ll|ll|ll}
\hline
\multirow{3}{*}{\begin{tabular}[c]{@{}c@{}}Temporal \\ Sampling \\ \end{tabular}} & \multicolumn{6}{c}{KoNViD-1k~\cite{hosu2017konstanz}}\\ \cline{2-7} & \multicolumn{2}{c|}{TSN}& \multicolumn{2}{c|}{TSM}& \multicolumn{2}{c}{ECO}\\ \cline{2-7} 
& \multicolumn{1}{c}{PLCC} & \multicolumn{1}{c|}{SRCC} & \multicolumn{1}{c}{PLCC} & \multicolumn{1}{c|}{SRCC} & \multicolumn{1}{c}{PLCC} & \multicolumn{1}{c}{SRCC} \\ \hline
OO& 0.72 & 0.73 & 0.72 & 0.72 & 0.73& 0.73                      \\ 
OS& 0.72 & 0.73 & 0.71 & 0.72 & 0.73 & 0.73                      \\ 
SO& 0.72 & \textbf{0.74} & \textbf{0.73}& \textbf{0.74} & 0.73 & 0.72              \\ 
SS& \textbf{0.73} & 0.73 & 0.72 & 0.73 & \textbf{0.74} & \textbf{0.74} \\ \hline

\multirow{3}{*}{\begin{tabular}[c]{@{}c@{}}Temporal \\ Sampling \\ \end{tabular}} & \multicolumn{6}{c}{LIVE-VQC~\cite{sinno2018large}}\\ \cline{2-7} & \multicolumn{2}{c|}{TSN}& \multicolumn{2}{c|}{TSM}& \multicolumn{2}{c}{ECO}\\ \cline{2-7} 
& \multicolumn{1}{c}{PLCC} & \multicolumn{1}{c|}{SRCC} & \multicolumn{1}{c}{PLCC} & \multicolumn{1}{c|}{SRCC} & \multicolumn{1}{c}{PLCC} & \multicolumn{1}{c}{SRCC} \\ \hline
OO& 0.69 & 0.62 & 0.70 & 0.66 & 0.69& 0.66                      \\ 
OS& 0.69 &0.64 & 0.69 & 0.63 & 0.70 & 0.66                      \\ 
SO& 0.70 & 0.64 & \textbf{0.71} & \textbf{0.67} & 0.68 & 0.64                       \\ 
SS& \textbf{0.70} & \textbf{0.66} & 0.70 & 0.66 & \textbf{0.72} & \textbf{0.67}\\ \hline
\end{tabular}}
\end{table}

\begin{figure*}
\centering
\subfigure[ Score: 3.53 ]{
\begin{minipage}[]{0.48\linewidth}
\includegraphics[width=1\linewidth]{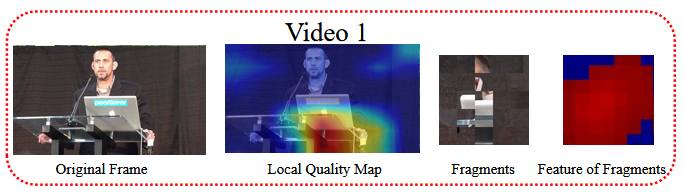}\vspace{0.51pt}
\end{minipage}
}
\subfigure[Score: 2.50 ]{
\begin{minipage}[]{0.48\linewidth}
\includegraphics[width=1\linewidth]{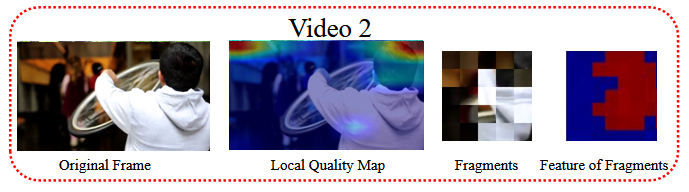}\vspace{0.51pt}
\end{minipage}
}
\subfigure[Score: 1.27 ]{
\begin{minipage}[]{0.48\linewidth}
\includegraphics[width=1\linewidth]{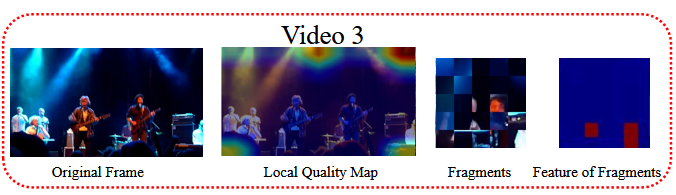}\vspace{0.51pt}
\end{minipage}
}
\subfigure[ Score: 3.48]{
\begin{minipage}[]{0.48\linewidth}
\includegraphics[width=1\linewidth]{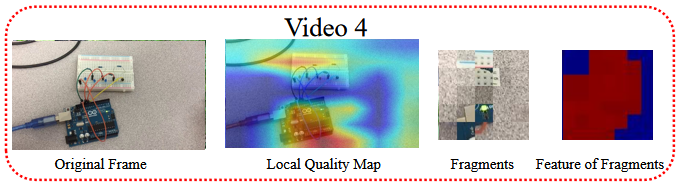}\vspace{0.51pt}
\end{minipage}
}
\subfigure[ Score: 2.9 ]{
\begin{minipage}[]{0.48\linewidth}
\includegraphics[width=1\linewidth]{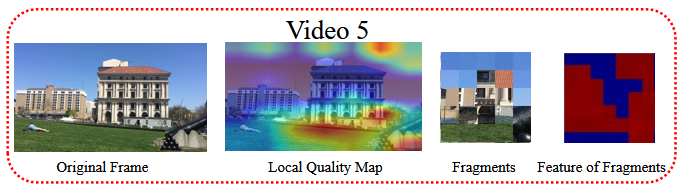}\vspace{0.51pt}
\end{minipage}
}
\subfigure[ Score: 1.96]{
\begin{minipage}[]{0.48\linewidth}
\includegraphics[width=1\linewidth]{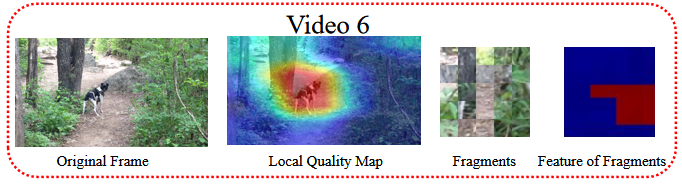}\vspace{0.51pt}
\end{minipage}
}

\caption{The visual examples of spatio-temporal local quality maps, where blue areas refer to relatively low quality scores and red areas refer to high scores. (a)(b)(c) are from the KoNViD-1k dataset with a resolution of 540p; (d)(e)(f) are from the LIVE-VQC dataset with a resolution of 1080p.}
\label{fig:vision}
\end{figure*}

Followed by our previous study~\cite{FangTIP2023}, which demonstrates that both ordered and orderless video frames are able to predict the associated video quality, we further test whether these ordered-or-orderless input formats are also suitable in current settings. Note that the experiments in this paper differ from our previous study~\cite{FangTIP2023} in that the temporal information in the extracted frames is lost due to the adopted random spatial patches extraction, while the input frames in~\cite{FangTIP2023} contain implicit or explicit temporal information since the frames are intact. Experiments are conducted on the KoNViD-1k and LIVE-VQC datasets with four input modes: OO (Order-Order), OS (Order-Shuffle), SO (Shuffle-Order), and SS (Shuffle-Shuffle), where the former means the input frame order in training phase and the latter denotes the input frame order in testing phase, respectively. The results, shown in Table \ref{tab:result9}, suggest that on the LIVE-VQC dataset, the model performance does not differ significantly with varied input sequences under three different temporal sampling conditions, which is complementary to the results in our previous study~\cite{FangTIP2023}.

\subsection{Qualitative Visualization}
The spatio-temporal local quality maps~\cite{wu2022fast} are generated to visualize what the model has learned. As shown in Figure~\ref{fig:vision}, six visual examples are selected from the KoNViD-1k and LIVE-VQC datasets, as well as their perspective local quality maps and segments with quality maps. For visual examples (a), (d), and (e), the proposed model focuses on regions of interest and assigns high quality to these regions, aligning with the perception mechanisms of the HVS. For example (b), the proposed model gives low quality to the blurry background, thus dropping its global quality. For example (c), a different pattern emerges, \ie, a dim and cluttered scene, where the proposed model successfully handles this case. And for example (f), the proposed model allocates high quality to the object regions, while the visual quality of most background areas is low, leading to a low global quality.

\section{Conclusion}
\label{sec:conc}

Our research aims to delve into the effectiveness-efficiency trade-offs issue of spatio-temporal modeling by immensely reducing video's information, which is implemented by joint spatial and temporal sampling. Through extensive experiments on six public datasets, we find that the heavily squeezed video can be also used to predict the quality of original video, and we further demonstrate the feasibility of pursuing an online VQA model. The experimental results of this study offer complementary insights on VQA to recent studies~\cite{yan2022subjective,FangTIP2023,sun2024analysis,wu2023neighbourhood}, and we expect that our study together with these studies~\cite{yan2022subjective,FangTIP2023,sun2024analysis,wu2023neighbourhood} will motivate more interesting work.

\ifCLASSOPTIONcaptionsoff
  \newpage
\fi

\bibliographystyle{IEEEtran}
\bibliography{egbib}


 

\begin{IEEEbiography}[{\includegraphics[width=1.1in,height=1.3in,clip,keepaspectratio]{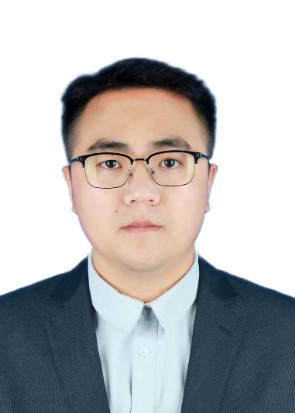}}]{Jiebin Yan} received the Ph.D. degree from Jiangxi University of Finance and Economics, Nanchang, China. He was a computer vision engineer with MTlab, Meitu. Inc, and a research intern with MOKU Laboratory, Alibaba Group. From 2021 to 2022, he was a visiting Ph.D. student with the Department of Electrical and Computer Engineering, University of Waterloo, Canada. He is currently a Lecturer with the School of Information Management, Jiangxi University of Finance and Economics, Nanchang, China. His research interests include visual quality assessment and computer vision.
\end{IEEEbiography}

\begin{IEEEbiography}[{\includegraphics[width=1.0in,height=1.3in,clip,keepaspectratio]{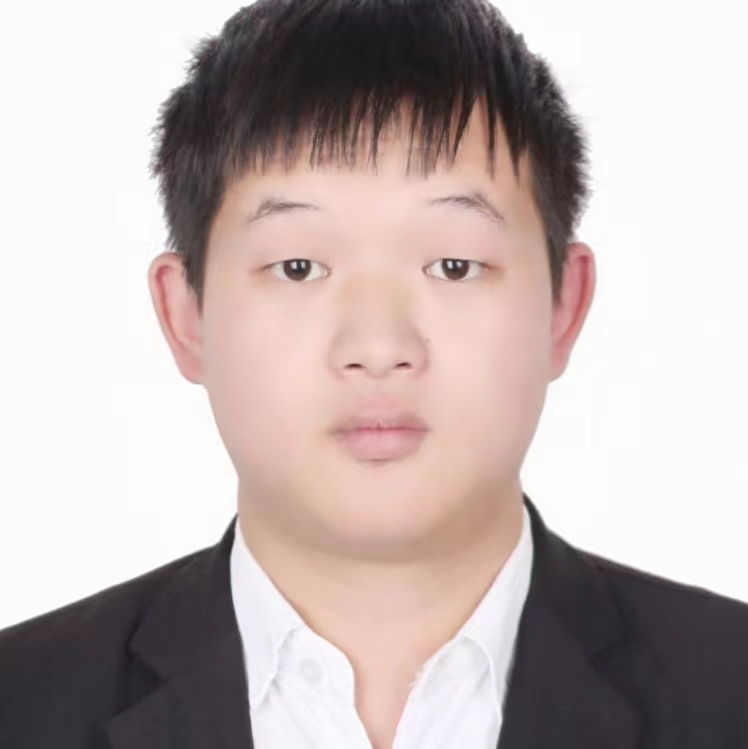}}]{Lei Wu} received his bachelor's degree from the School of Software of Nanchang Hangkong University in 2022, and he is pursuing his master's degree with the School of Information Management of Jiangxi University of Finance and Economics. His research interests include video/panoramic video quality assessment and computer vision.
\end{IEEEbiography}

\begin{IEEEbiography}[{\includegraphics[width=1.0in,height=1.3in,clip]{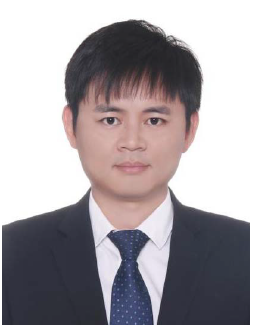}}]{Yuming Fang}(M’13-SM’17) received the B.E. degree from Sichuan University, Chengdu, China, the M.S. degree from the Beijing University of Technology, Beijing, China, and the Ph.D. degree from Nanyang Technological University, Singapore. He is currently a Professor with the School of Information Management, Jiangxi University of Finance and Economics, Nanchang, China. His research interests include visual attention modeling, visual quality assessment, computer vision, and 3D image/video processing. He serves on the editorial board for IEEE Transactions on Multimedia and Signal Processing: Image Communication.
\end{IEEEbiography}

\begin{IEEEbiography}[{\includegraphics[width=1.0in,height=1.3in,clip]{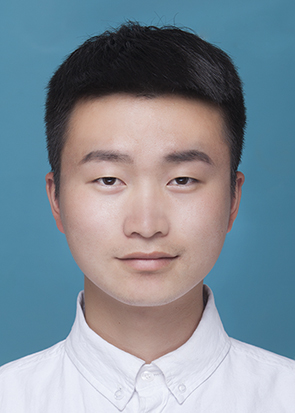}}]{Xuelin Liu} received the B.E, M.S, and Ph. D. degrees from Jiangxi University of Finance and Economics, Nanchang, China. He is currently a postdoctoral fellow with Jiangxi University of Finance and Economics. From 2023 to 2024, he was a visiting PhD student with the Department of Computer Science, City University of Hong Kong. His research interests include visual quality assessment and computer vision.
\end{IEEEbiography}

\begin{IEEEbiography}[{\includegraphics[width=1.0in,height=1.3in,clip]{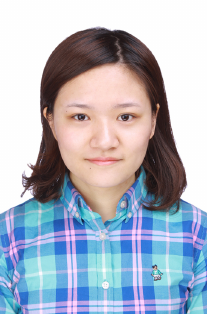}}]{Xue Xia} received the B.E. and M.E. degrees from Shanghai University, Shanghai, China, and the D.E. degree from the Jiangxi University of Finance and Economics, Nanchang, China. She is currently a lecturer with the School of Information Management, Jiangxi University of Finance and Economics, Nanchang, China. Her research interests include pattern recognition, medical image processing and semantic segmentation.
\end{IEEEbiography}

\begin{IEEEbiography}[{\includegraphics[width=1.0in,height=1.2in,clip]{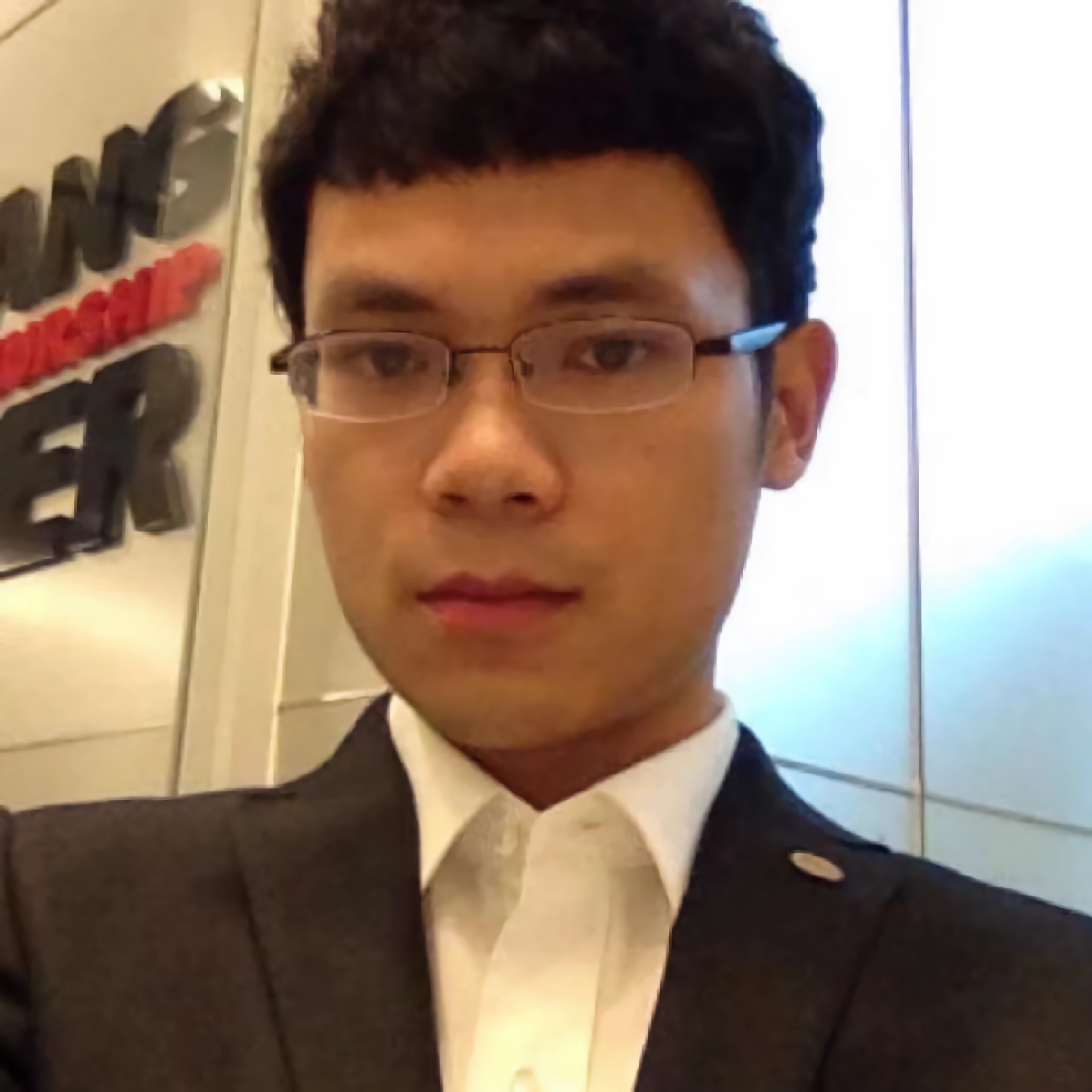}}]{Weide Liu}  is currently a Research Fellow at Harvard Medical School, Harvard University. Before that, he was a Research Scientist at A*STAR, Singapore. Weide received his Ph.D. from Nanyang Technological Unversity. His research interests include computer vision, language, machine learning, federated learning, and medical image analysis.
\end{IEEEbiography}

\end{document}